\documentclass[10pt,twocolumn,letterpaper]{article}

\usepackage{wacv}              

\usepackage{graphicx}
\usepackage{amsmath}
\usepackage{amssymb}
\usepackage{booktabs}
\usepackage{pgf-pie}
\usepackage{tikz}
\usepackage{svg}
\usepackage{array}
\usepackage{adjustbox}
\usepackage{multirow}
\usepackage{arydshln}
\usepackage{xcolor}        
\usepackage{placeins}

\usetikzlibrary{positioning,chains,calc,fit,backgrounds,arrows.meta, shapes.geometric,shadows,shadows.blur,spy,matrix}
\colorlet{matteC}{blue!65!black}
\definecolor{fluxC}{HTML}{994D4E}
\definecolor{scaleC}{HTML}{3F7C3E}

\tikzset{
  >=Stealth,
  stage/.style={draw, thick, rounded corners=10pt,
                minimum width=2.2cm, minimum height=1.2cm,
                align=center, font=\bfseries},
  innerbox/.style={
        draw, thick, rounded corners=4pt,
        align=center, font=\bfseries,
        minimum width=1.6cm,   
        minimum height=.9cm,
        text width=2.0cm,
        inner sep=0pt},
  picstage/.style 2 args={
    stage,
    inner sep=0pt,
    path picture={
      \node[inner sep=0pt] at (path picture bounding box.center)
        {\includegraphics[
        width=\pgfkeysvalueof{/pgf/minimum width},
        height=\pgfkeysvalueof{/pgf/minimum height},
        keepaspectratio]{#1}};},
    label=center:{#2}},
  step/.style={draw, thick, rounded corners=4pt, font=\scriptsize,
               text width=3.6cm, align=center,
               minimum height=0.6cm, inner sep=3pt},
  decision/.style={draw, diamond, aspect=2, thick,
                   inner sep=0pt, font=\scriptsize, align=center},
  arrFlux/.style  ={->,thick,color=fluxC},
  arrMatte/.style ={->,thick,color=matteC},
  arrScale/.style ={->,thick,color=scaleC},
  arrDefault/.style={->,thick},
  hex/.style={
  draw, thick, shape=regular polygon, regular polygon sides=6,
  inner sep=1pt, align=center, font=\bfseries, minimum width=2.6cm
    },
  side/.style={font=\scriptsize, align=center},
  side_box/.style={draw, thick, rounded corners=6pt,
                minimum width=2.2cm, minimum height=0.6cm,font=\scriptsize, align=center},
  vsep/.style={dashed, gray, line width=.4pt},
  node distance = 16mm and 22mm,
  start chain=going right,
  every on chain/.style={join=by arrDefault},
  dbl dashed/.style={
    dash pattern=on 4pt off 3pt,      
    line width=1pt,
    postaction={draw,
      dash pattern=on 4pt off 3pt,
      line width=1pt,
      yshift=50pt                      
    }}            
}

\definecolor{wacvblue}{rgb}{0.21,0.49,0.74}
\usepackage[pagebackref,breaklinks,colorlinks,allcolors=wacvblue]{hyperref}
\usepackage[capitalize]{cleveref}

\def\wacvPaperID{446} 
\def\confName{WACV}
\def\confYear{2026}
\newcommand{\modelname}{Odo} 
\newcommand{\datasetname}{ChangeLing18K}
\newcommand{\trek}{\raisebox{-0.6ex}{\includegraphics[height=1.6em]{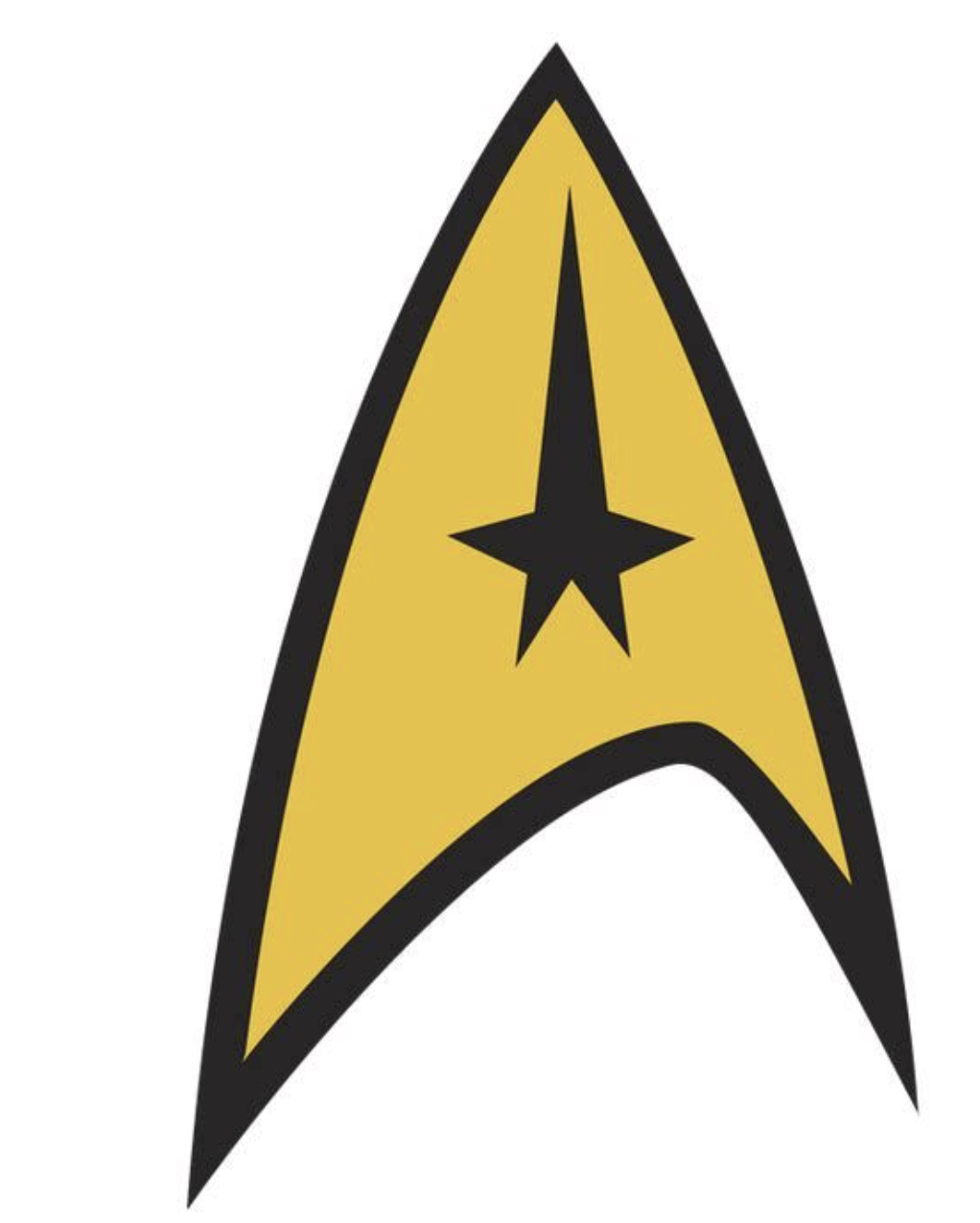}}}

\title{Odo\trek: Depth-Guided Diffusion for Identity-Preserving Body Reshaping}

\author{
Siddharth Khandelwal \quad Sridhar Kamath \quad Arjun Jain \\
Fast Code AI Consult Pvt. Ltd. \\
{\tt\small siddharth@fastcode.ai \quad sridhar@fastcode.ai \quad arjunjain@gmail.com}
}

\begin{document}
\twocolumn[{
    \maketitle
    \begin{center}
        \captionsetup{type=figure}
        \includegraphics[width=\linewidth]{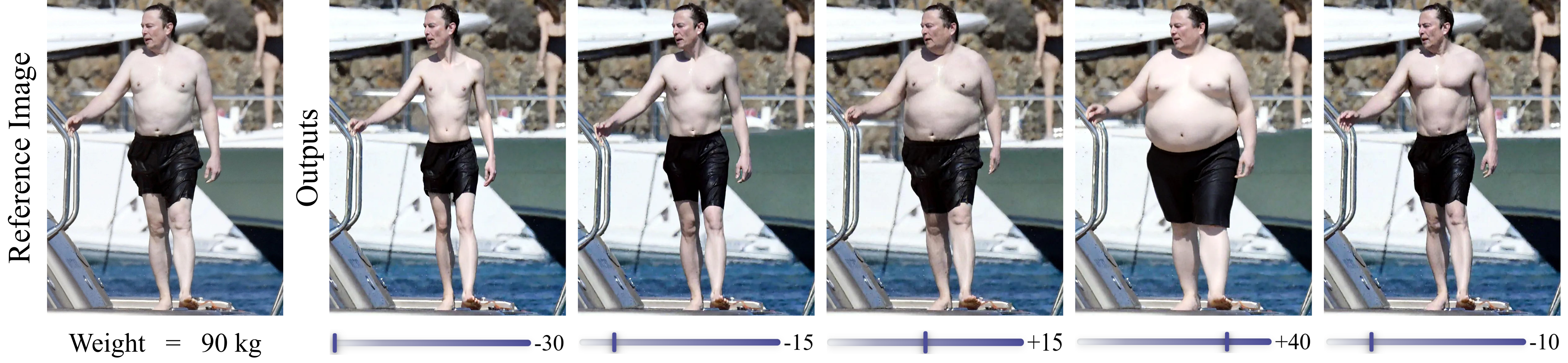}
        \captionof{figure}{
        \textbf{\modelname}~enables realistic body shape transformation by adjusting semantic sliders (e.g., weight) that directly modify the SMPL model. The modified SMPL then guides the transformation of the input image. Natural language prompts like “make the person muscular,” “fatter,” or “thinner” further refine the output. In addition to the weight parameter, in the last column we make the person muscular by giving the prompt ``Make the person muscular''. The model supports accurate shape edits from a single reference image.}
        \label{fig:variations}
    \end{center}
}]
\begin{abstract}
Human shape editing enables controllable transformation of a person's body shape, such as thin, muscular, or overweight, while preserving pose, identity, clothing, and background. Unlike human pose editing, which has advanced rapidly, shape editing remains relatively underexplored. Current approaches typically rely on 3D morphable models or image warping, often introducing unrealistic body proportions, texture distortions, and background inconsistencies due to alignment errors and deformations. A key limitation is the lack of large-scale, publicly available datasets for training and evaluating body shape manipulation methods.
In this work, we introduce the first large-scale dataset of 18,573 images across 1523 subjects, specifically designed for controlled human shape editing. It features diverse variations in body shape, including fat, muscular and thin, captured under consistent identity, clothing, and background conditions. Using this dataset, we propose \modelname, an end-to-end diffusion-based method that enables realistic and intutive body reshaping guided by simple semantic attributes. Our approach combines a frozen UNet that preserves fine-grained appearance and background details from the input image with a ControlNet that guides shape transformation using target SMPL depth maps. Extensive experiments demonstrate that our method outperforms prior approaches, achieving per-vertex reconstruction errors as low as 7.5mm, significantly lower than the 13.6mm observed in baseline methods, while producing realistic results that accurately match the desired target shapes.

\end{abstract}
    
\section{Introduction}
\label{sec:intro}

Digital human shape editing is a critical task, enabling realistic manipulation of body shapes across various domains, such as entertainment, virtual fashion, health visualization, and personalized digital avatars.  Moreover, it can support medical interventions by visualizing healthy body shapes for conditions such as anorexia and bulimia nervosa~\cite{psych1,psych2}. Human shape editing involves modifying a person's body size, shape, and proportions in images, making them appear thinner, heavier, or more muscular while preserving key attributes like pose, identity, clothing, and background.

Previous methods for human shape editing have relied on 3D morphable models or image warping techniques, which often produce visual artifacts and background distortions under large deformations due to their dependence on explicit geometric correspondences. While human pose editing has advanced significantly with large-scale datasets~\cite{liuLQWTcvpr16DeepFashion}, gathering data of the same person with multiple distinct body shapes presents significant practical challenges. Unlike poses, which can be easily captured in short sessions or extracted from videos, body shape variations are difficult to obtain. This absence of large-scale datasets with identity-consistent body shape variations remains a critical barrier to developing robust shape editing techniques

\begin{figure}[!t]
  \centering
  \includegraphics[width=\linewidth]{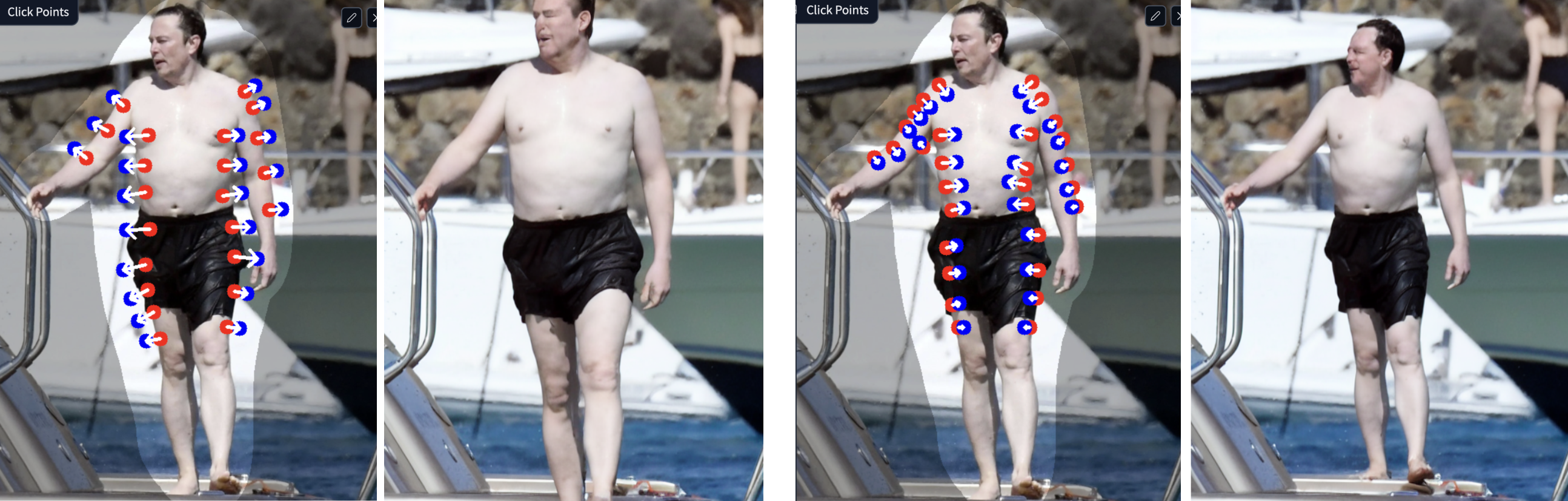}
  \caption{DragDiffusion~\cite{Shi_2024_CVPR} for fat and thin transformations}
  \label{fig:drag_results}
\end{figure}

To address this, we introduce \textbf{\datasetname}, the first large-scale dataset specifically curated for human shape editing. It features diverse body types including thin, muscular, overweight, with each pair depicting the same subject across different shapes while maintaining consistent identity, pose, clothing, and viewpoint. This facilitates precise and identity-preserving model training.
Building on this dataset, we propose \textbf{\modelname}, to the best of our knowledge, the first ever end-to-end diffusion-based framework designed specifically for human shape editing.
Recently, diffusion models have demonstrated remarkable success in different image-editing tasks~\cite{9880056,Chung_2024_CVPR,lu2024coarse,Kim_2024_CVPR}. Their iterative denoising mechanism, combined with strong priors learned from large datasets, enables fine-grained control and high-quality outputs. However, traditional methods like text-based or point-dragging-based methods~\cite{brooks2022instructpix2pix,Shi_2024_CVPR} are not capable of human body shape transformation as shown in~\cref{fig:drag_results}. These techniques lack the precise control for editing human body shapes often causing artifacts or undesired results.

\modelname~achieves precise, linear, and scalable shape control through the integration of explicit shape guidance into the diffusion process through a combination of a frozen SDXL UNet~\cite{podell2024sdxl} and a depth ControlNet-conditioned~\cite{Zhang_2023_ICCV} diffusion model. The frozen UNet preserves identity, appearance, and background details from the input image, while SMPL-based depth maps condition the diffusion model towards achieving accurate and visually coherent target body shapes. Users can interactively adjust semantic body attributes through intuitive sliders, facilitating realistic and precise reshaping without compromising visual fidelity.

We also introduce the first comprehensive benchmark specifically designed to quantitatively evaluate human shape editing performance. Our benchmark provides metrics assessing both image quality relative to ground truth and the accuracy of target body shape transformations. Experimental results demonstrate that \modelname~outperforms existing methods across all evaluated metrics by a large margin.

In summary, our contributions are threefold: (1) \datasetname, the first large-scale dataset specifically designed for human shape editing; (2) \modelname, a novel diffusion-based model capable of realistic and precise shape transformations trained on our dataset; and (3) a comprehensive benchmark for quantitatively evaluating human shape editing. We will publicly release our dataset and model to promote future research and reproducibility in human body reshaping.


\section{Related Work}

In this section we discuss the evolution of human-shape editing methods, highlight general diffusion-based techniques, and the current scope of SMPL conditioning in diffusion models.\\

\noindent\textbf{Human Shape Editing.}
Human body shape editing has evolved through several approaches, each with distinct advantages and limitations.

Zhou~et~al.~\cite{zhou2010parametric} pioneered parametric body reshaping by fitting 3D morphable body model (SCAPE~\cite{scape}) via user-guided pose alignment and contour-based shape fitting. Semantic edits applied to the fitted model are transferred back to the image through body-aware warping. MovieReshape~\cite{movie-reshape} by Jain~et~al. extended this by automating the fitting process, enabling temporally coherent reshaping in videos. Richter~et~al.~\cite{richter} introduced a real-time reshaping system using Kinect sensors, employing automatic shape initialization, pose tracking, segmentation refinement, and foreground-specific image warping. Despite their effectiveness, these 3D model-based approaches require manual intervention and generalize poorly beyond constrained datasets. They also suffer from fitting inaccuracies propagating into noticeable warping artifacts, pose tracking instabilities under occlusions, and also unintended global warping effects on closely positioned body parts.

Ren~et~al.~\cite{ren2022structure} introduced a method that predicts pixel-level deformation flows guided by body structure using skeletons and Part Affinity Fields. Their Structure Affinity Self-Attention ensures consistent changes across related body parts by leveraging visual and structural relationships. However, this approach struggles with significant shape edits, causing unnatural warping and background distortions.

Okuyama~et~al.~\cite{Okuyama_2024_WACV} combine SMPL-X parametric body model initialization with diffusion-based refinement for pose and shape editing. They fit and modify the 3D model, followed by a two-stage diffusion refinement with weak noise to preserve identity and remove projection artifacts. However, their approach requires per-subject fine-tuning and careful noise schedule tuning. It also struggles with loose clothing and background distortions due to projection of the original image onto the edited SMPL model. Consequently, these methods remain ineffective for robust human body reshaping, often resulting in unrealistic outputs with background distortions.

\noindent\textbf{Image Editing using Diffusion.}
Foundational work such as Denoising Diffusion Probabilistic Models (DDPM)~\cite{ho2020denoising} and Latent Diffusion Models (LDM)~\cite{Rombach_2022_CVPR} introduced scalable frameworks that enable controllable and high-resolution image synthesis. These models have served as the foundation for numerous diffusion-based image-editing works across tasks including inpainting, colorization, semantic modification, and subject personalization.

Methods such as Palette~\cite{10.1145/3528233.3530757} and SDEdit~\cite{meng2022sdedit} have demonstrated the capability of diffusion models for image-to-image tasks like colorization and sketch-to-image synthesis. Prompt-to-Prompt~\cite{hertz2022prompt} provided fine-grained attention control, enabling localized edits without retraining. Building upon these techniques like, Imagic~\cite{kawar2023imagic} and InstructPix2Pix~\cite{brooks2022instructpix2pix} introduced semantically rich, text-driven image editing. Null-text inversion~\cite{mokady2022null} proposed an efficient inversion scheme for real images by optimizing the unconditional textual embedding used in classifier-free guidance, enabling accurate reconstruction and intuitive prompt-driven editing without fine-tuning the model weights. Meanwhile, works like DragDiffusion~\cite{Shi_2024_CVPR} and DragonDiffusion~\cite{mou2024dragondiffusion} enabled point-based image editing by allowing users to manipulate images by dragging handle points to target points.

These methods are designed for semantic, structural, or stylistic editing but are unsuitable for large-scale anatomical transformations in the human body while preserving fine-grained identity and clothing details~\cref{fig:drag_results}. Human shape editing requires manipulating body proportions in a controllable manner, beyond the capacity of latent inversion or prompt-conditioned editing.

\noindent\textbf{SMPL Conditioned Diffusion.}
Recent methods integrate SMPL models into diffusion frameworks to enhance spatial and anatomical consistency in human image generation. CHAMP~\cite{zhu2024champ} utilizes SMPL-derived spatial representations as conditioning inputs to a latent diffusion model. By aligning these SMPL conditions with a reference image's shape parameters, CHAMP achieves temporally coherent human animations. However, its scope is limited to pose variation and does not support  body shape editing. Buchheim~et~al.~\cite{buchheim2025controlling} propose an SMPL-conditioned ControlNet for text-to-image generation, encoding SMPL shape and pose parameters into the diffusion UNet cross-attention. However, their method requires synthetic-to-real domain adaptation and does not facilitate shape editing. These methods show the potential of SMPL conditioning to enhance human shape control within diffusion models.
\begin{figure}[!t]  
  \centering
  \setlength{\tabcolsep}{0.5pt}  
  \renewcommand{\arraystretch}{1}

  \begin{tabular}{@{}*{3}{>{\centering\arraybackslash}m{0.33\linewidth}}@{}}   
    \scriptsize Thin & \scriptsize Fat & \scriptsize Muscular \\[2pt]
    \includegraphics[height=3.5cm, keepaspectratio]{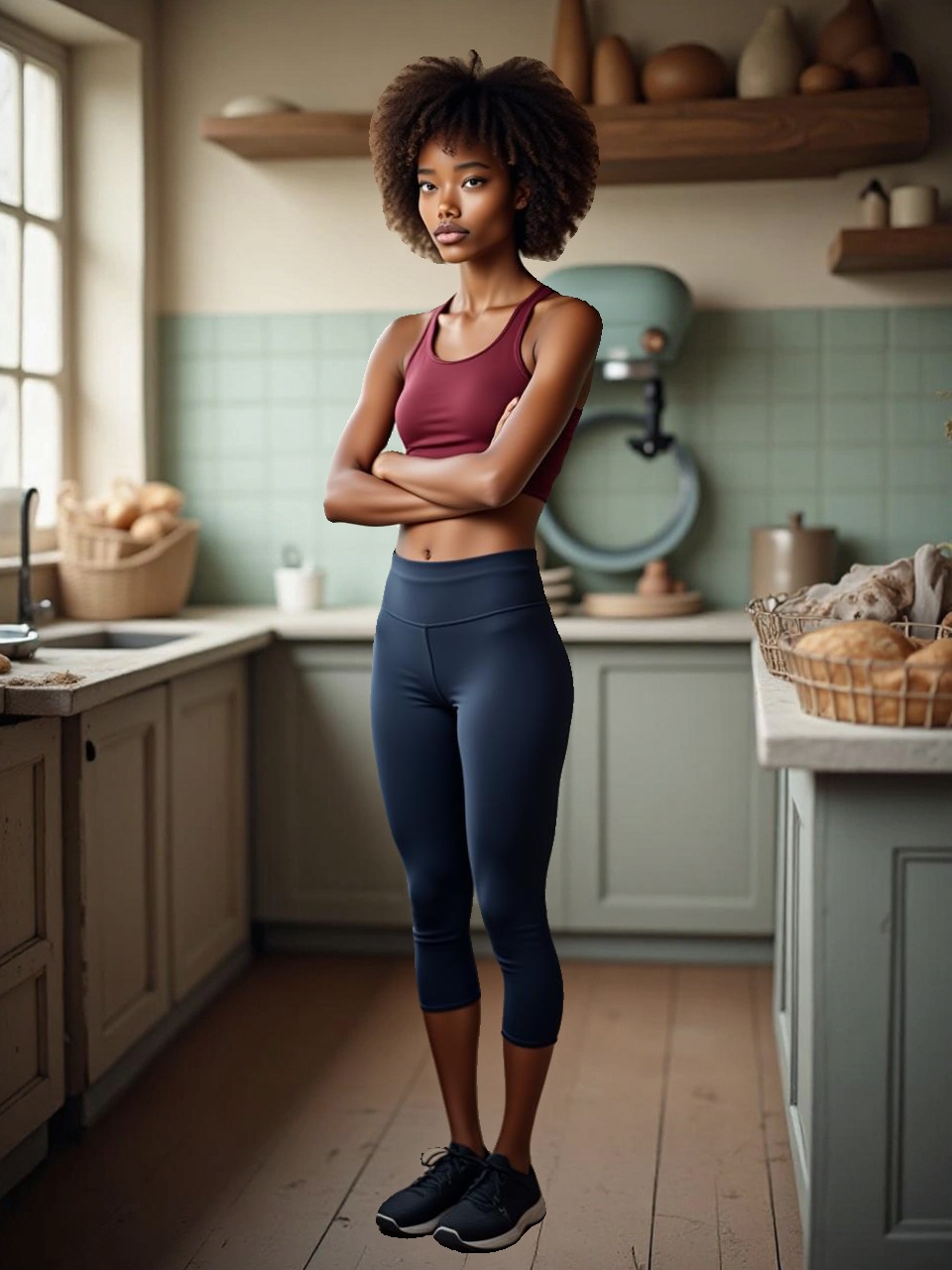} &
    \includegraphics[height=3.5cm, keepaspectratio]{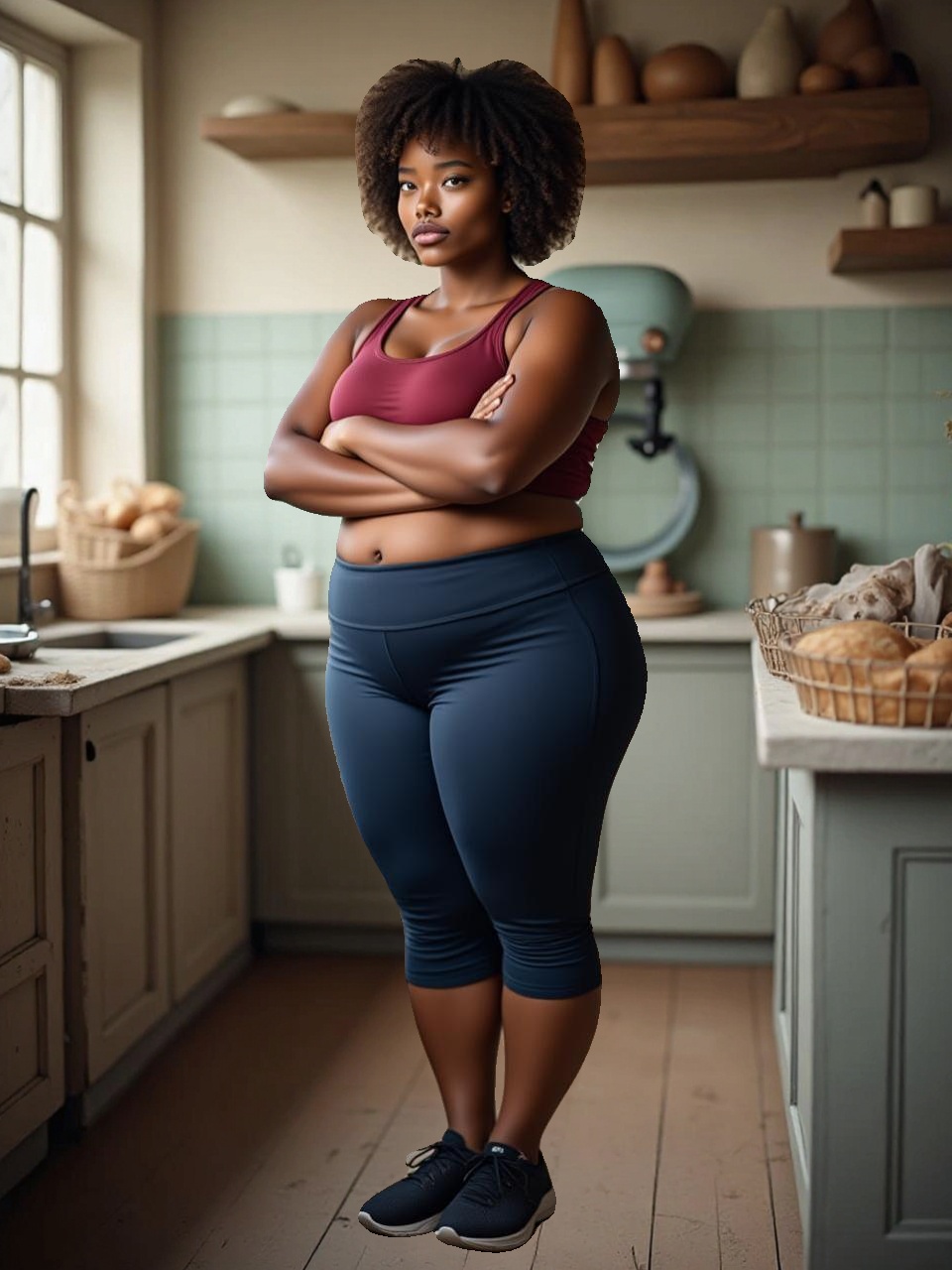} &
    \includegraphics[height=3.5cm, keepaspectratio]{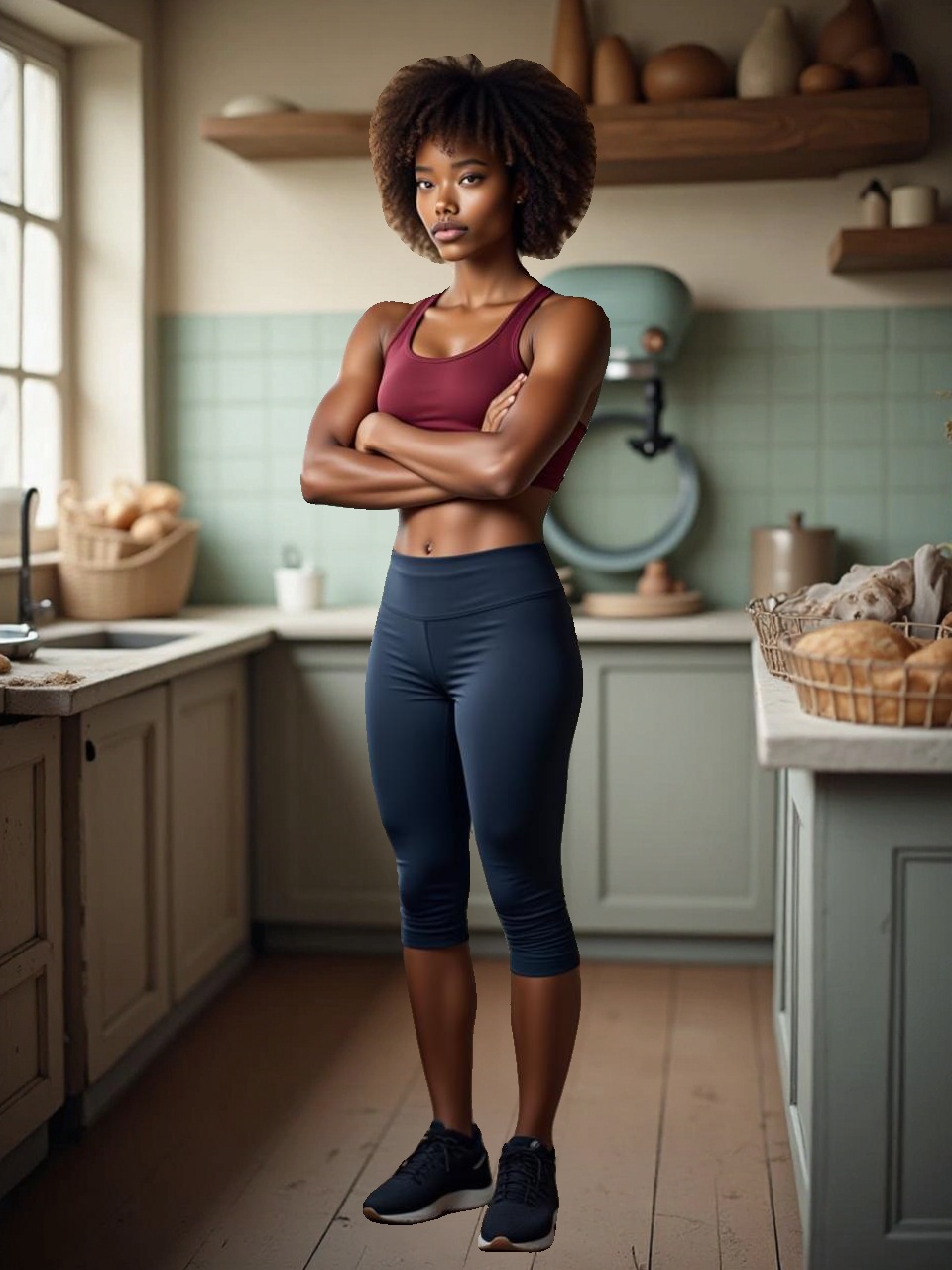}\\[-1.5pt]
    \includegraphics[height=3.5cm, keepaspectratio]{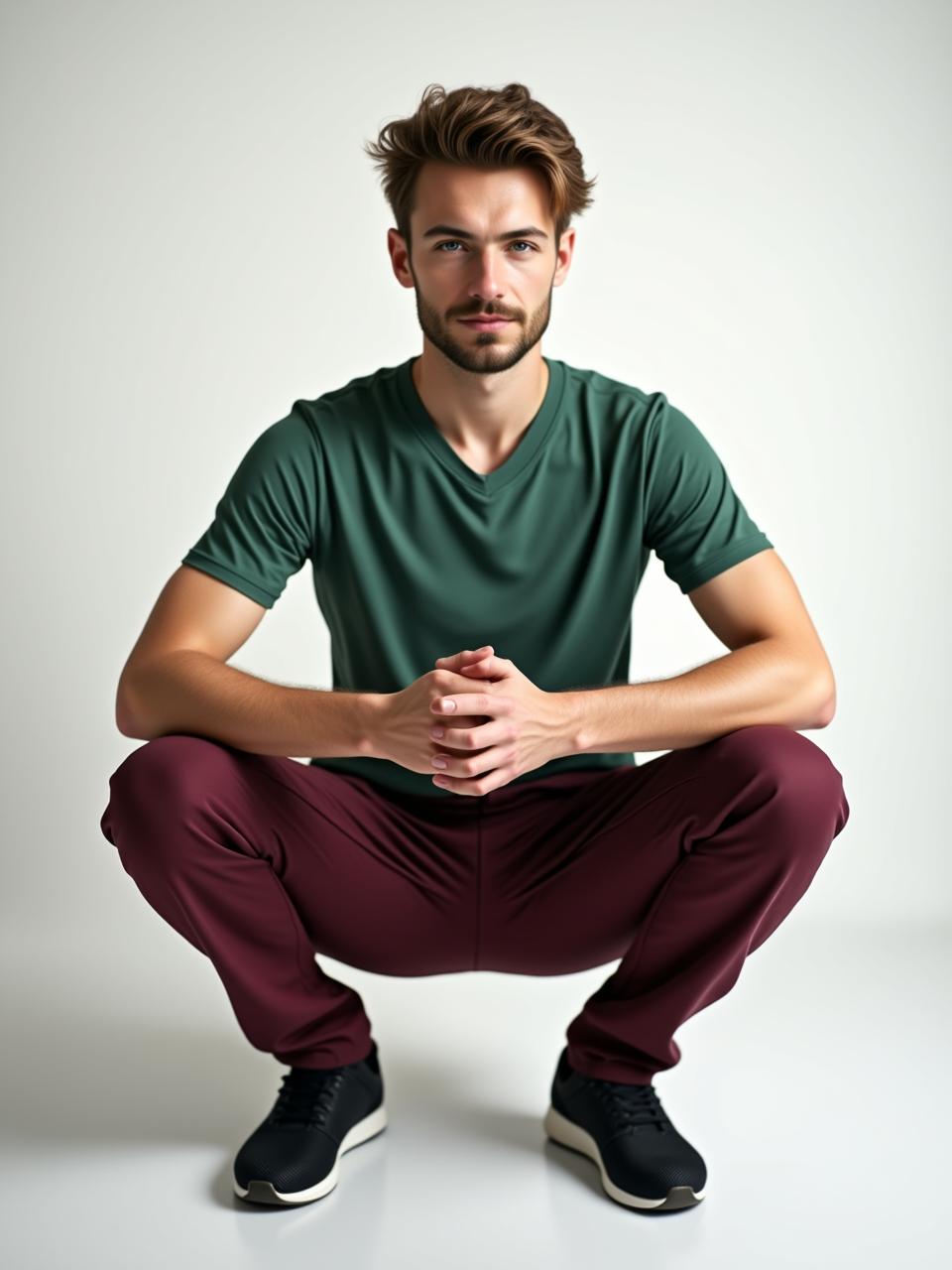} &
    \includegraphics[height=3.5cm, keepaspectratio]{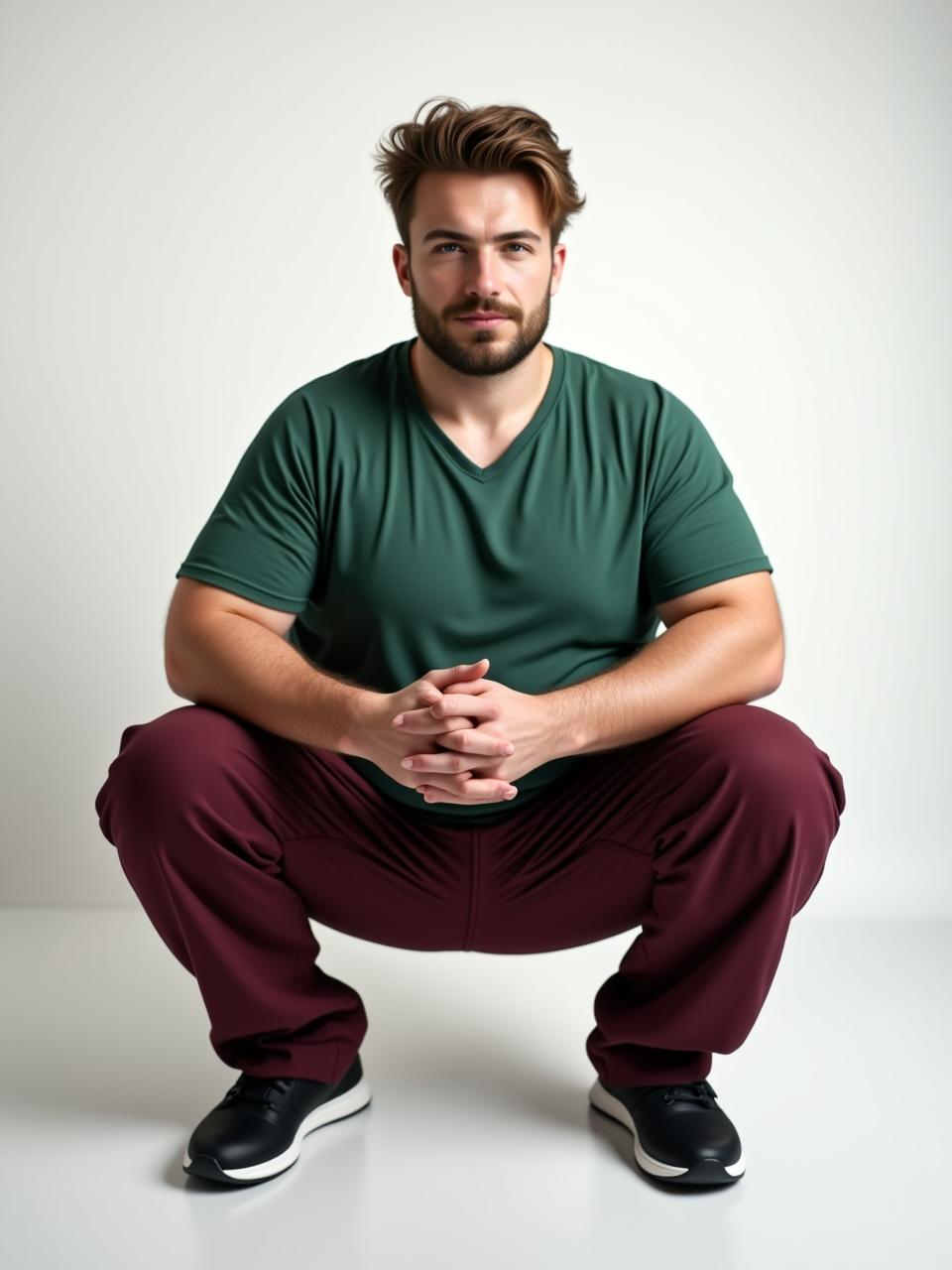} &
    \includegraphics[height=3.5cm, keepaspectratio]{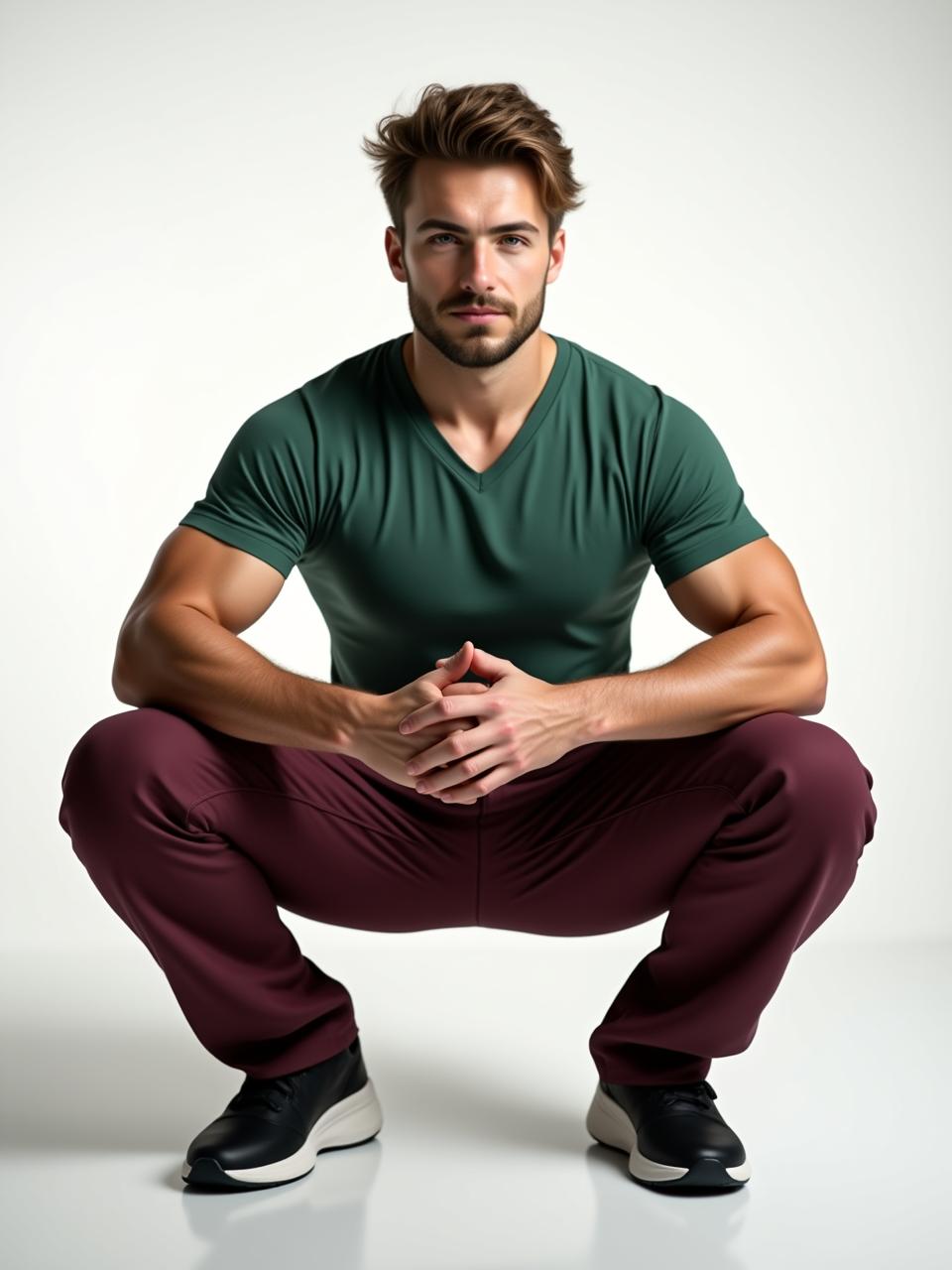}\\[-1.5pt]
  \end{tabular}
  \caption{ Example dataset images showcasing subjects in thin, fat, and muscular forms. The dataset includes such images with diverse backgrounds, poses, clothing, and body transformations.}
  \label{fig:dataset_images}
\end{figure}

\section{\datasetname~Dataset}
\label{sec:dataset}
Despite substantial progress in image generation, large-scale, high-resolution datasets tailored specifically for human body shape editing remain scarce. Existing datasets typically lack controlled shape variations of the same individual under consistent identity, pose, clothing, and background conditions. For instance, the BR-5K dataset~\cite{ren2022structure} provides 5,000 high-quality portrait pairs with professionally retouched versions; however, each image has the face-blurred out and includes only a single minimal transformation, which is often barely noticeable (see~\Cref{tab:dataset_comparison}). To address this gap, we introduce \datasetname{}, a novel large-scale dataset specifically designed for identity-consistent human shape transformations.

\begin{table}[!h]
    \centering
    \caption{Comparison between BR-5K and Our Body Shape Transformation Dataset. Qualitative comparison is provided in the supplementary material.}
    \label{tab:dataset_comparison}
    \small{\begin{tabular}{@{}lcc@{}}
        \toprule
        \textbf{Feature} & \textbf{BR-5K} & \textbf{Our Dataset} \\ 
        \midrule
        \# of Pairs & 5,000 & 18.573 \\ 
        Body Shape Variation Extent & Minimal & Wide  \\ 
        Identity Consistency & - & Yes \\ 
        \% of Female Pairs & 92.4 & 44.2 \\ 
        \% of Male Pairs & 7.6 & 55.8 \\
        \bottomrule
    \end{tabular}}
\end{table}

\begin{figure*}[!t]
\centering
  \resizebox{\textwidth}{!}{\input{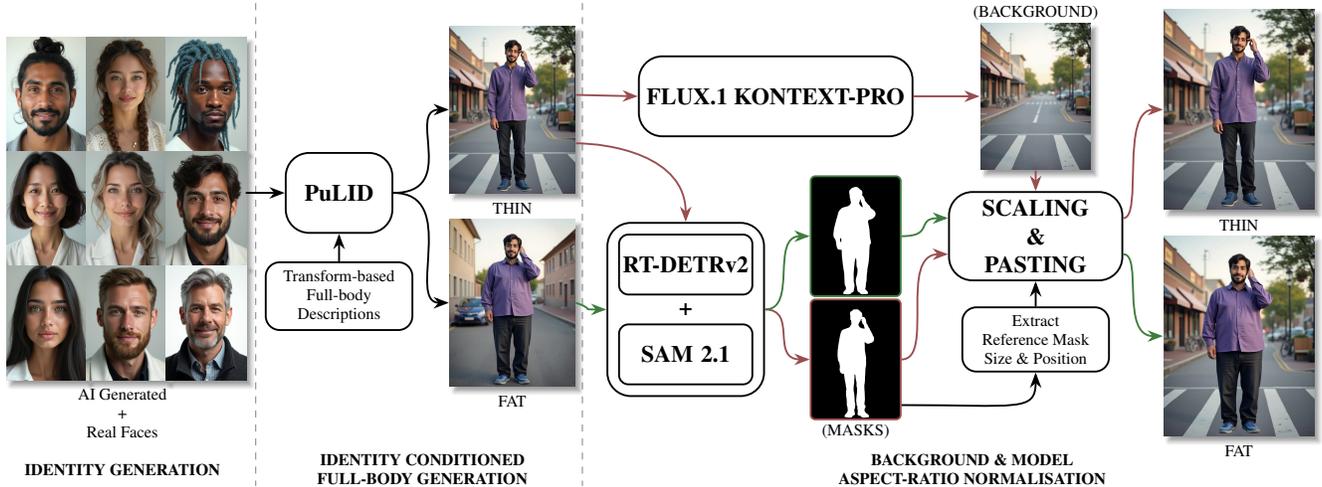}} 
  \caption{End-to-end dataset generation pipeline. Identities are created using FLUX.1-dev\cite{flux2024} and license-free face images. PuLID\cite{guo2024pulid}, a diffusion-based tuning-free ID customization method for text-to-image generation, generates thin, fat, and muscular variations for each identity. Human masks from RT-DETRv2\cite{lv2024rtdetrv2improvedbaselinebagoffreebies}+SAM2.1\cite{ravi2025sam} are scaled and pasted onto a common FLUX-Kontext\cite{labs2025flux1kontextflowmatching} refined background for consistency.}
  \label{fig:dataset_pipeline}
\end{figure*}

\subsection{Dataset Generation Workflow}

\datasetname~is synthesized through a four–stage, model–driven pipeline that guarantees \emph{identity fidelity}, \emph{pose and attire consistency}, and \emph{high–resolution realism} across three body–types (\textit{thin}, \textit{fat}, \textit{muscular}) \cref{fig:dataset_images}. The dataset generation pipeline is shown in \cref{fig:dataset_pipeline} and detailed below.\\

\noindent\textbf{1.~Identity Creation:}
We initialize each synthetic subject with a high–resolution facial portrait generated by \textbf{FLUX.1-dev}, a 12-B parameter rectified–flow diffusion transformer that excels at photorealistic face synthesis from textual prompts~\cite{flux2024}. To increase demographic diversity, we additionally sample license–free reference faces from Pexels~\cite{pexels} and Unsplash~\cite{unsplash}. In total, we create a corpus of 1523 human face images. \\

\noindent\textbf{2.~Identity-conditioned Full–body Generation:}
Given the face portrait, we utilize \textbf{PuLID}~\cite{guo2024pulid} checkpoint built upon the \textsc{FLUX.1}-dev base model. PuLID’s contrastive ID loss helps preserve facial likeness across generations. We supply the model with (i) the facial image and (ii) a standardized prompt specifying \emph{gender}, \emph{clothing}, \emph{pose}, \emph{scene}, and \emph{body type} (\textit{thin, fat, muscular}). For each face, we generate 5 prompt sets, where each set consists of three prompts (thin, fat, muscle) differing only in the \emph{body type} description. This results in 15 generations per face (5 per body type), yielding 7,615, high-resolution (960\(\times\)1280) images per body type, differing exclusively in body shape. \\

\begin{figure*}[!t]
  \centering
  \includegraphics[width=\linewidth]{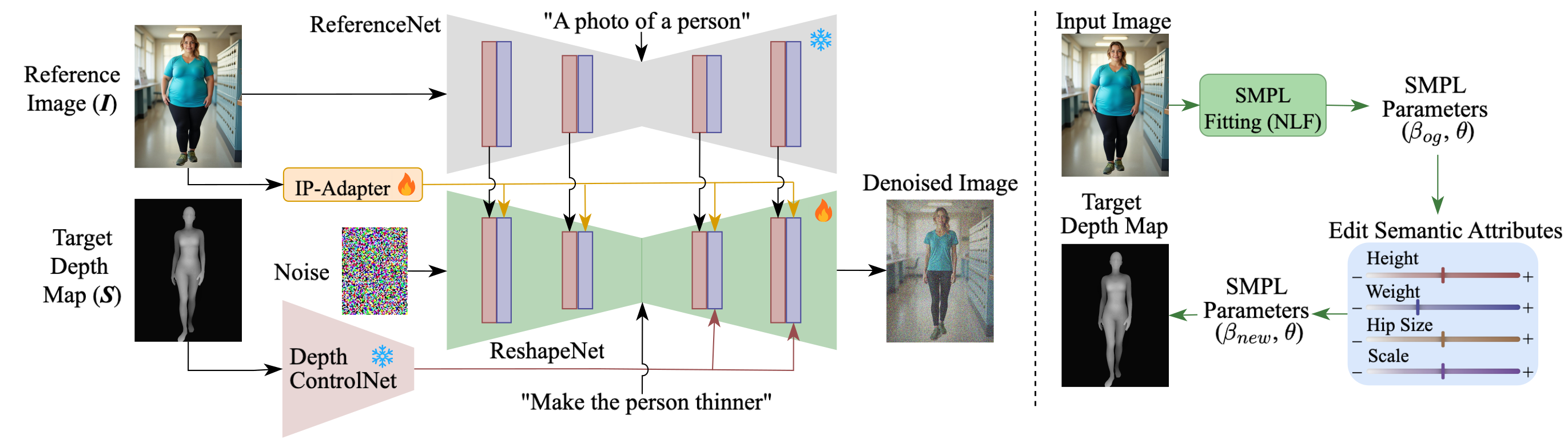}
  \caption{The left figure illustrates our training pipeline, where the target SMPL depth-map (\textbf{S}) derived from the target image conditions the ReshapeNet via a Depth ControlNet for the desired shape transformation. Features from the reference image, extracted by the ReferenceNet, are integrated into the ReshapeNet using spatial self-attention. The right figure depicts the inference pipeline, where SMPL parameters are initialized from the input image, adjusted based on semantic attributes, and rendered into the target SMPL depth map. This depth map, along with input image features, conditions the ReshapeNet during inference to generate the final output}
  \label{fig:pipeline}
\end{figure*}

\noindent\textbf{3.~Background and Aspect–ratio Normalization:}
The three body type images for each identity sometimes exhibit slight shifts in background alignment and minor differences in perceived subject size. This variation arises due to the inherent stochastic nature of diffusion models: each generation begins from a new Gaussian noise seed, and even subtle alterations in the input prompt—such as changing only the body type description—can influence the model's latent-space trajectories, resulting in these visual discrepancies. To eliminate this variability we carry out a lightweight, fully automatic post-process: \\\\
\textbf{i) Reference Image selection.} The \textit{thin} image is treated as the baseline reference because it's smaller silhoutte exposes the most background, yielding a near complete scene for subsequent background completion step. \\
\textbf{ii) Subject Segmentation.} We first detect the person in the reference image using \emph{RT-DETRv2}\cite{lv2024rtdetrv2improvedbaselinebagoffreebies}, then pass the resulting bounding box to \emph{SAM 2.1}\cite{ravi2025sam} to produce a binary mask. The same detector–segmenter pipeline is used to obtain masks for the remaining two body types. \\
\textbf{iii) Background Completion.} We pass the \textit{thin} reference image to \emph{FLUX.1 Kontext [pro]}~\cite{labs2025flux1kontextflowmatching} prompting it to remove the person and complete the background. This yields a clean background image shared by the entire triplet (thin, fat, muscle). \\
\textbf{iv) Uniform Scaling \& Composition.} From the thin reference mask we measure the height and record the bottom-most coordinates of the mask. Each non-thin mask is resized uniformly to match it's height, maintaining aspect ratio, and composited onto the clean background at the same coordinates. This ensures consistent framing across images while preserving real shape variations.

The resulting transformation triplets (thin, fat and muscle) have an identical background and uniform subject scale. This removes irrelevant variations that could negatively affect subsequent training or evaluation.\\

\noindent\textbf{4.~Manual Curation.}
Each matched triplet (\textit{thin}, \textit{fat}, \textit{muscular}) allows forming six transformation pairs: \textit{thin→fat}, \textit{thin→muscular}, and \textit{fat→muscular} and their reverse orders, giving a theoretical total of 7,615\(\times\)6\(=\)45,690 transformation pairs. After manual filtering out pairs with mismatched clothing, large pose discrepancies, unnatural poses or limb geometry, limb duplication, warped garments, identity drift, or negligible shape change, 18,573 high-quality transformation pairs remain for training and evaluation.

While minor pose variations exist across shape transformations for each identity, our model remains robust to these differences.
\section{Method}
Given the above dataset we propose \textbf{\modelname}, a diffusion-based approach for reshaping humans. Each pair in the dataset constitutes an input reference image and a target image of a person captured with consistent clothing and background, but a different body shape. \modelname~incorporates a target SMPL depth map derived from the target image to act as shape conditioning during training. The SMPL model~\cite{SMPL:2015} represents the human body's 3D shape with parameters $\boldsymbol{\beta} \in \mathbb{R}^{10}$ and pose with joint rotation parameters $\boldsymbol{\theta} \in \mathbb{R}^{72}$. To generate the target depth maps, we first fit a SMPL model to the target image using the method introduced in~\cite{sarandi2024nlf}. The optimized $\boldsymbol{\beta}$ and $\boldsymbol{\theta}$ parameters are subsequently used to render spatially aligned depth maps.

Given an input image $\boldsymbol{I}$ and the target SMPL depth map $\boldsymbol{S}$, our objective is to transform the person in $\boldsymbol{I}$ to match the shape in $\boldsymbol{S}$. 

For inference on in-the-wild images without target SMPL depth maps, we use a mapping between semantic attributes and SMPL $\boldsymbol{\beta}$
 shape parameters~\cite{Gralnik_2023_ICCV}. This mapping enables modification of semantic attributes like weight, height, and body size to control the SMPL $\boldsymbol{\beta}$ shape parameters. Adjusting these parameters modifies the depth map, ensuring the generated output image aligns with the desired body shape.

\subsection{Model Architecture}
Our model~(\cref{fig:pipeline}) consists of a ReshapeNet supported by three key modules: a ReferenceNet, an IP-Adapter~\cite{ye2023ip-adapter}, and a Depth ControlNet~\cite{Zhang_2023_ICCV}. The ReferenceNet captures detailed features (background, clothing, identity) from input image $\boldsymbol{I}$ and feeds them to the ReshapeNet. The IP-Adapter provides high-level feature assistance, while the Depth ControlNet performs SMPL conditioning for body shape transformation. We discuss each module and their role in more detail below. \\

\noindent\textbf{ReferenceNet.}
To retain fine details of the input image, the IP-Adapter alone is insufficient. Following previous works~\cite{hu2023animateanyone, xu2023magicanimate, choi2024improving, zhu2024champ}, we incorporate a frozen SDXL UNet to extract intermediate image features. These intermediate latents are injected into the self-attention blocks of the ReshapeNet by concatenating them with the hidden states before performing spatial self-attention. This operation is formulated as:
\begin{equation}
    \text{Attn}(\mathbf{Q'}, \mathbf{K'}, \mathbf{V'}) = \text{Softmax}\left(\frac{\mathbf{Q'} \mathbf{K'}^T}{\sqrt{d}}\right) \mathbf{V'}
\label{eq: spatial_attention}
\end{equation}
where $\mathbf{Q'} = \mathbf{W_q}[z, y]$, $\mathbf{K'} = \mathbf{W_k}[z, y]$, and $\mathbf{V'} = \mathbf{W_v}[z, y]$. Here, $z$ denotes the intermediate latents of the ReshapeNet, $y$ represents the corresponding intermediate latents from the frozen UNet encoder, and $[z, y]$ indicates concatenation. We pass only the first half of the attention output to subsequent layers. The ReferenceNet uses a fixed prompt ``A photo of a person'' throughout training. \\

\noindent\textbf{Image Prompt Adapter.}
To condition the diffusion model with high-level features from the input image, we employ the IP-Adapter module~\cite{ye2023ip-adapter}. The IP-Adapter encodes the input image using a CLIP image encoder, then projects the features through linear and self-attention layers to generate image embeddings. These embeddings are integrated into the ReshapeNet via cross-attention, analogous to text embeddings. The combined cross-attention operation is:
\begin{equation}
   Z^{\text{new}} = \text{Attn}(\mathbf{Q}, \mathbf{K_t}, \mathbf{V_t}) + \text{Attn}(\mathbf{Q}, \mathbf{K_i}, \mathbf{V_i})
  \label{eq:ip-adapter}
\end{equation}
where $\mathbf{Q}$ is the query matrix from the ReshapeNet's intermediate representation, $\mathbf{K_t}$ and $\mathbf{V_t}$ are the key and value matrices from text embeddings, and $\mathbf{K_i}$ and $\mathbf{V_i}$ are the key and value matrices from image embeddings. \\

\noindent\textbf{SMPL Depth ControlNet.} ControlNets~\cite{Zhang_2023_ICCV} enable conditioning of text-to-image diffusion models using spatial inputs such as Canny edges, normal maps, and depth maps. In \modelname, we integrate a Depth ControlNet that applies conditioning to the middle and decoder blocks of the ReshapeNet through residual connections. The conditioning depth map is rendered from the target SMPL parameters and is aligned with the target image. \\

\noindent\textbf{ReshapeNet.}
The ReshapeNet serves as the base UNet in our model, built upon the SDXL UNet architecture~\cite{podell2024sdxl}. During training, the target image is encoded through a variational autoencoder to obtain latents $\boldsymbol{z_0}$. Noise is progressively added to produce $\boldsymbol{z_t}$ at timestep $t$, which serves as input to the ReshapeNet along with features from the ControlNet and ReferenceNet. The UNet output is then decoded to produce the denoised image.

To guide category-specific transformations, we incorporate targeted text prompts based on the transformation type: ``Make the person fatter'', ``Make the person thinner'', or ``Make the person muscular''. These prompts provide crucial semantic control, particularly for muscularity transformations where SMPL-derived depth maps are insufficient. While depth maps effectively capture coarse shape attributes like overall body size and proportions, they cannot encode fine-grained features such as muscle definition. Prompt-based conditioning compensates for this limitation by introducing high-level semantic intent, enhancing the model's ability to generate accurate and detailed body modifications. We demonstrate the effectiveness of these prompts in~\cref{sec:ablations}.

\subsection{Training Setup}
\label{sec:training-setup}
Given an input reference image $\boldsymbol{I}$, target SMPL depth map $\boldsymbol{S}$, and text prompt $\boldsymbol{c_t}$, the model learns to predict noise added to the noised latent $\boldsymbol{z_t}$ using the standard diffusion learning objective:
\begin{equation}
    \mathcal{L} = \mathbb{E}_{\boldsymbol{z_0},t,\boldsymbol{c_t},\boldsymbol{I},\boldsymbol{S},\epsilon\sim\mathcal{N}(0,1)} \left[ \| \epsilon - \epsilon_{\theta}(\boldsymbol{z_t},t,\boldsymbol{c_t},\boldsymbol{I},\boldsymbol{S}) \|_2^2 \right]
\end{equation}
where $\epsilon_{\theta}$ is the model's noise prediction parameterized by trainable parameters $\theta$.
\section{Experiments}
We conduct comprehensive experiments to evaluate our approach, including qualitative and quantitative comparisons with existing methods across diverse test cases, along with ablation studies to validate key design choices.

\subsection{Implementation Details}
We train \modelname~on our training set with all SMPL depth maps precomputed to minimize memory usage and enhance training efficiency. Training is conducted on a single NVIDIA A100 GPU (80GB VRAM) for 60 epochs using mixed precision to improve memory efficiency and training speed. Images are resized to 768$\times$1024. We use the Adam optimizer~\cite{kingma2014adam} with learning rate $1 \times 10^{-5}$, $\beta_1 = 0.9$, $\beta_2 = 0.999$, weight decay $1 \times 10^{-2}$, and $\epsilon = 1 \times 10^{-4}$. For initialization, the ReshapeNet uses pre-trained SDXL UNet weights and is fine-tuned jointly with the IP-Adapter (initialized from its pre-trained checkpoint). The ReferenceNet is initialized with SDXL weights and kept frozen during training. The ControlNet uses pre-trained depth ControlNet weights~\cite{xinsir2023controlnetdepth} and also remains frozen throughout training. Our model requires approximately 23GB GPU memory and 18 seconds for single-image inference.

\begin{figure}[!t]  
  \centering
  \includegraphics[width=\columnwidth]{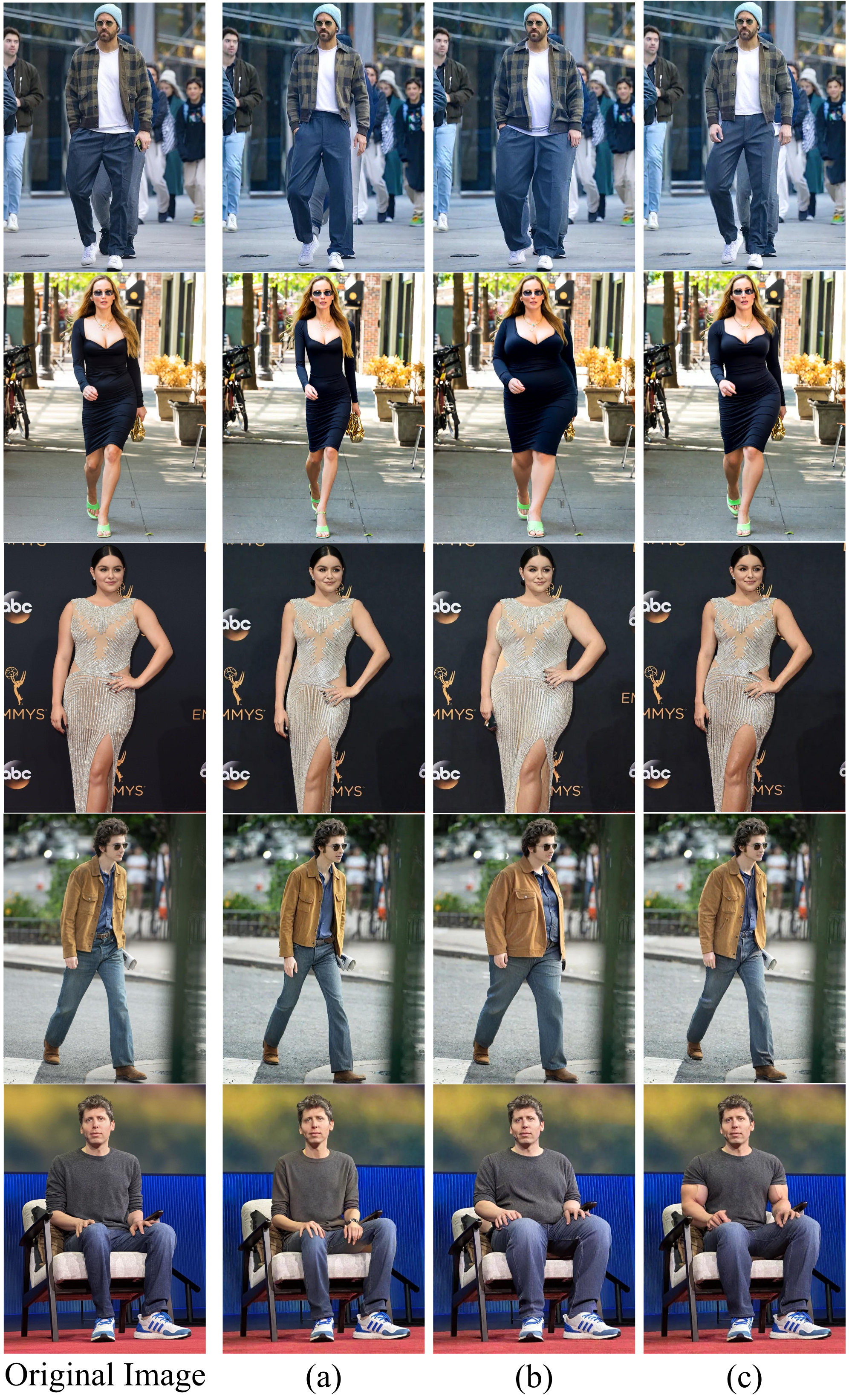}
  \vspace{-2em}
\caption{\modelname~generates realistic body transformations while preserving identity, clothing, and background details. Examples show transformations from original images to (a) thinner, (b) overweight, and (c) muscular body types across various poses including sitting and standing positions.}
  \label{fig:in-the-wild}
\end{figure}

\subsection{Evaluation}
As noted by Okuyama~et~al.~\cite{Okuyama_2024_WACV}, no dataset exists for objective quantitative evaluation of body shape editing with large deformations. We therefore construct the first benchmark dataset for evaluating human body shape editing, consisting of 3,600 image pairs. This benchmark dataset is constructed using the exact same pipeline as the training data but uses distinct real face images and background descriptions to prevent training data leakage. It includes diverse body shape variations for a comprehensive assessment.

We evaluate generated images against the ground truth using four key metrics. For image fidelity, we use Structural Similarity Index Measure (\textbf{SSIM})~\cite{Wang2004SSIM}, Peak Signal-to-Noise Ratio (\textbf{PSNR}) for pixel-level accuracy, and Learned Perceptual Image Patch Similarity (\textbf{LPIPS})~\cite{zhang2018unreasonable} for perceptual similarity. For body shape accuracy, we fit SMPL models to both ground truth and generated images using~\cite{sarandi2024nlf}, then compute the \textbf{S}cale \textbf{C}orrected \textbf{P}er-\textbf{V}ertex \textbf{E}uclidean error in neutral (\textbf{T}-)pose (\textbf{PVE-T-SC})~\cite{STRAPS2020BMVC}. This metric quantifies how accurately the generated image preserves the target body shape.

\begin{figure}[!t]
\centering
  \resizebox{\columnwidth}{!}{\input{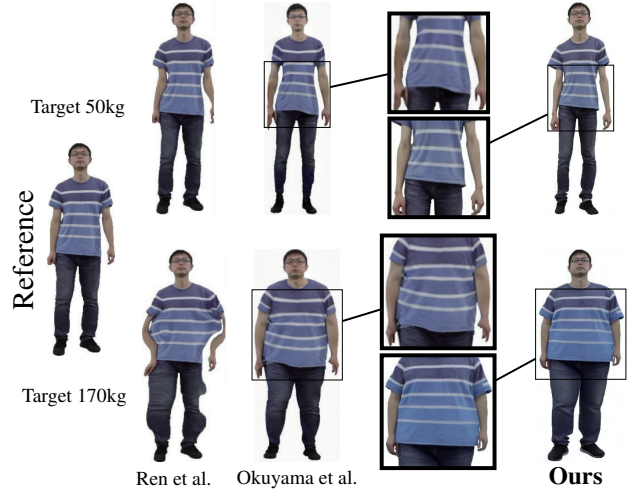}}
\caption{Qualitative comparison of body shape editing for target weights of 50kg and 170kg. Compared to Ren~et~al.~\cite{ren2022structure} and Okuyama~et~al.~\cite{Okuyama_2024_WACV}, our method produces more anatomically consistent body shapes and realistic clothing behavior, including proportionally adjusted hands, smoother fabric flow, better crease placement, and properly aligned shirt stripes.}
  \label{fig:qualitative_comp}
\end{figure}

\noindent\textbf{In-the-wild Images.}
We perform an out-of-distribution testing of our model by performing edits on in-the-wild images. \cref{fig:in-the-wild} demonstrates this, where our method effectively handles diverse poses, backgrounds, and clothing while preserving person identity. In addition to SMPL target shapes, we provide textual prompts—``Make the person fatter,'' ``Make the person thinner,'' or ``Make the person muscular''—to explicitly guide the desired transformations. We have provided more examples in the supplementary material.
\cref{fig:variations} further demonstrates our model's capability to perform diverse shape transformations. The model accurately follows SMPL depth maps to generate multiple variations of thinner and fatter versions from the reference image.

For a quantitative out-of-distribution evaluation we construct a separate dataset using Seedream 4.0~\cite{seedream4}. We collected real world images of people and created fatter, thinnner and muscular versions of them using the Seedream model. Although these edits are not  prefect and have some small variations it still serves as a good benchmark to analyse how faithfully \modelname~is able to follow target SMPL shapes. This dataset consists of 428 image pairs. Our model achieves a low PVE-T-SC of \textbf{11.69 mm} along with an SSIM of 0.67, PSNR of 18.81, and LPIPS of 0.25. In comparison, the method by Ren et al. achieves an SSIM of 0.62, PSNR of 18.47, LPIPS of 0.26, and PVE-T-SC of 13.82 mm, demonstrating our method's superior performance.


\noindent\textbf{Qualitative Comparisons.}
\cref{fig:qualitative_comp} presents a qualitative comparison of our approach with previous state-of-the-art methods by Ren~et~al.~\cite{ren2022structure} and Okuyama~et~al.~\cite{Okuyama_2024_WACV}. The former controls transformation through warping strength while the latter uses input weight and height. For fair comparison, we select target beta parameters such that the final SMPL-derived weight matches that shown by Okuyama~et~al.~\cite{Okuyama_2024_WACV}. Our results demonstrate more realistic transformations according to the target weight, as our model simultaneously adjusts overall body shape, limb proportions, and clothing, resulting in anatomically consistent and visually convincing modifications.

\noindent\textbf{Quantitative Comparisons.}
\cref{table:quantitative_comparison} presents evaluation metrics on our benchmark dataset. Our model outperforms all other methods across all metrics, achieving a PVE-T-SC of 7.52 mm, indicating that our generated body shapes closely match ground-truth shapes.

We compare against Ren~et~al.~\cite{ren2022structure}, setting warping strength according to the weight difference between input and ground truth images for fair evaluation. Since the implementation of Okuyama~et~al.~\cite{Okuyama_2024_WACV} is unavailable, quantitative comparison is limited to Ren~et~al.

We also evaluate FLUX.1 Kontext[dev]~\cite{flux12024kontext}, a prompt-based editing model. We design specific prompts instructing the model to \textit{``Make the person fatter,'' ``Make the person thinner,'' or ``Make the person muscular''} while specifying target weights. However FLUX.1 Kontext[dev] achieves limited performance, as shown in \cref{table:quantitative_comparison}.

\begin{table}[!t]
    \centering
\caption{Comparison of our model~\modelname~against Ren~et~al.~\cite{ren2022structure} and FLUX.1 Kontext[dev] on our test set. We also show results for three ablations: without prompt conditioning in ReshapeNet, without ReferenceNet, and training only on BR-5K data.}
    \label{table:quantitative_comparison}
    \small{
    \begin{tabular}{@{}lcccc@{}}  
        \toprule
        \textbf{Method} & \textbf{SSIM$\uparrow$} & \textbf{PSNR$\uparrow$} & \textbf{LPIPS$\downarrow$} & \textbf{PVE-T-SC$\downarrow$}\\ 
        \midrule
        Ren~et~al.~\cite{ren2022structure} & 0.6790 & 17.4567 & 0.2363 & 13.6337 \\
        FLUX.1 & 0.6788 & 16.5195 & 0.2826 & 19.1911 \\
        Kontext[dev]~\cite{flux12024kontext} & & & & \\
        Ours & \textbf{0.7716} & \textbf{19.0714} & \textbf{0.2035} & \textbf{7.4902} \\
        \hdashline \\[-1.8ex] 
        Ours w/o & 0.7281 & 15.1745 & 0.2566 & 9.8032 \\
        prompts &  &  &  & \\
        Ours w/o  & 0.6035 & 13.9982 & 0.4625 & 9.4157 \\
        ReferenceNet & & & & \\
        Ours with & 0.6939 & 14.1581 & 0.3432 & 18.6143 \\
        BR-5K data & & & & \\
        \bottomrule
    \end{tabular}}
\end{table}

\subsection{Ablations}
\label{sec:ablations}

We conduct ablation experiments to evaluate the effectiveness of various modules in our model. We investigate different methods for inputting image features into ReshapeNet and explore different prompting techniques to guide the model in specific shape transformation tasks. The quantitative results for these ablations are shown in \cref{table:quantitative_comparison}.

\begin{figure}[!t]  
  \centering
  \includegraphics[width=\columnwidth]{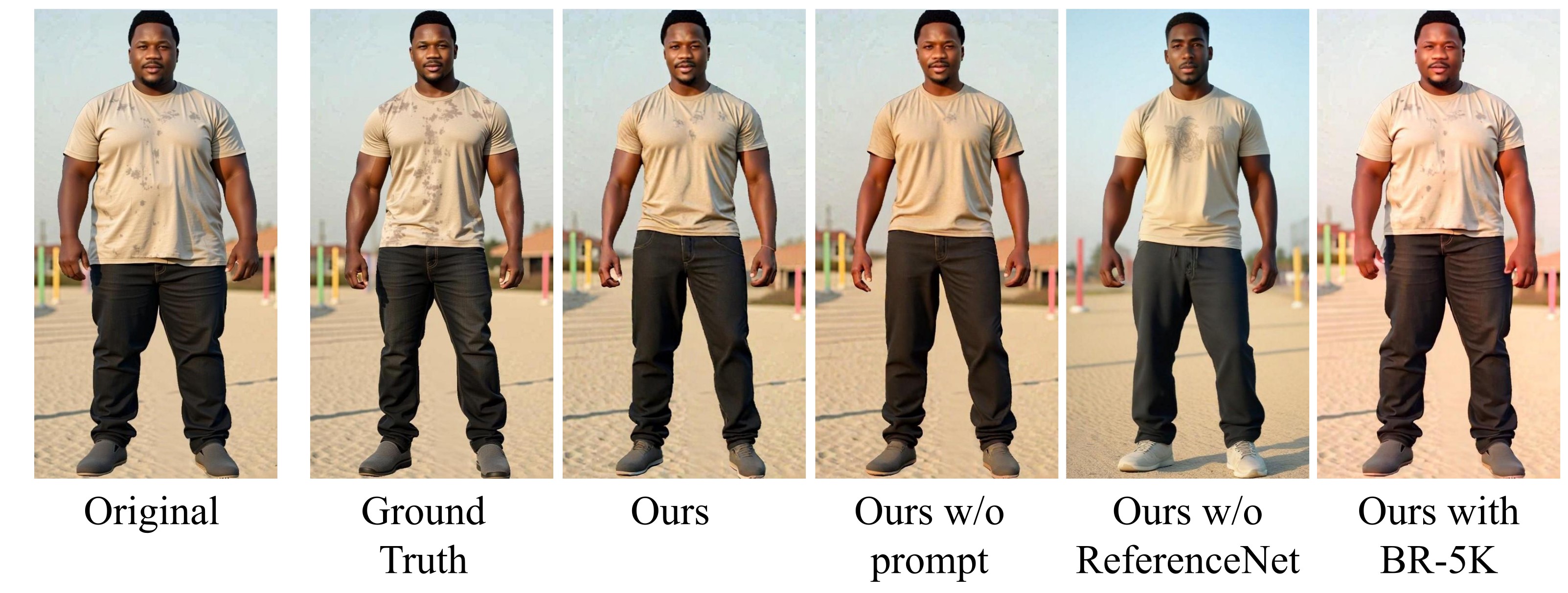}
  \vspace{-2em}
  \caption{Omitting the category-specific prompts preserves identity, clothing, and background but fails to generate muscular transformations, despite aligning with the SMPL depth map. Removing the ReferenceNet allows the intended shape transformation but, loses identity and background details. With BR-5K data, our model is unable to perform any visible shape tranformation.}
  \label{fig:ablations}
\end{figure}

We evaluate our method without the ReferenceNet by removing it and relying solely on a trainable IP-Adapter to incorporate input image features into the denoising UNet. As shown in \cref{fig:ablations}, omitting the ReferenceNet results in poor preservation of the background, attire (shoes in this example) and identity from the input image.

We also experiment with alternative prompts for the main denoising UNet. Specifically, instead of using category-specific prompts, we input a generic prompt such as ``A photo of a person'' for every pair during training. If only given this prompt during inference, the model remains effective at aligning with the SMPL depth maps and preserving clothing, background, and identity but struggles to enhance muscularity as shown in \cref{fig:ablations}.

We train our model only on the BR-5K dataset and evaluate it on our benchmark data to highlight the significance of our new dataset. Since the BR-5K dataset includes only very slight variations in body shape, the model is unable to learn to perform significant body shape transformations as seen in \cref{fig:ablations}. This is reflected by the high value of 18.61mm of PVE-T-SC obtained on the test set.

\section{Conclusion}
We introduce~\modelname, the first end-to-end diffusion-based framework for photorealistic, identity-preserving body shape editing at high resolution. Our approach generalizes across diverse images with varying poses and backgrounds by combining a ReferenceNet with a Depth ControlNet-conditioned diffusion model for fine-grained body shape modifications. We also introduce~\datasetname, a large-scale dataset containing 18,573 paired images with varied shape transformations while maintaining consistent identity, clothing, and backgrounds.

\begin{figure}[!h]  
  \centering
  \includegraphics[width=\columnwidth]{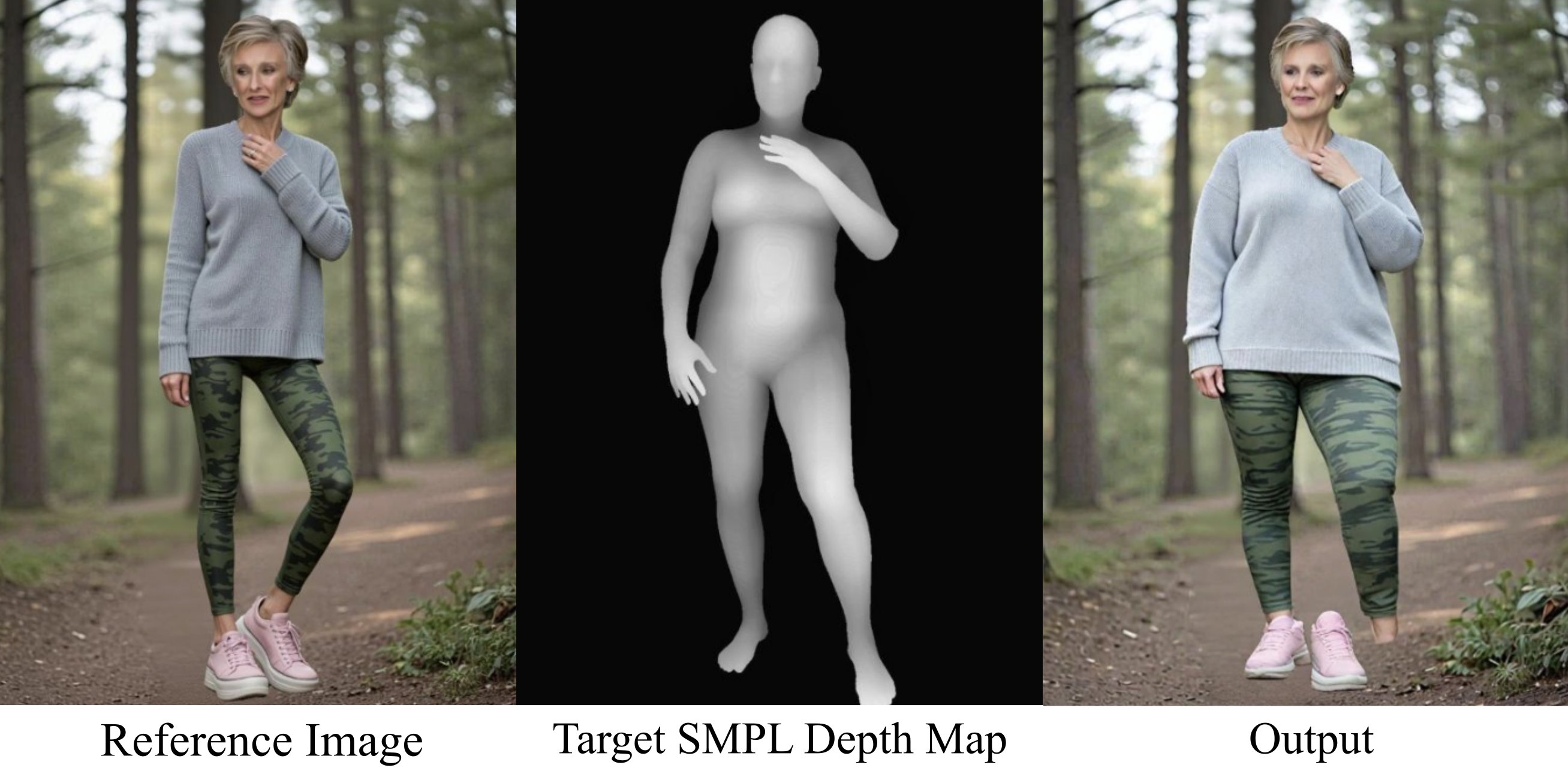}
  \vspace{-2em}
  \caption{Failure Case: The target SMPL depth map pose defers significantly from the reference image pose. This may lead to our model hallucinating the original pose in the output}
  \label{fig:failure_cases}
\end{figure}

While our approach demonstrates strong performance, there remain areas for improvement. Fine facial details can occasionally be challenging to preserve, leading to facial artifacts. When SMPL target poses differ significantly from input poses, the model may generate inconsistent body parts, as shown in~\cref{fig:failure_cases}. Fine clothing details (like text) can also be sometimes difficult to reproduce accurately. These limitations present opportunities for future work to enhance fidelity and robustness of human body reshaping.
{
    \small
    \bibliographystyle{ieeenat_fullname}
    \bibliography{main}
}

\end{document}